\documentclass[fleqn,9pt]{SelfArx} 
\usepackage{lipsum} 
\usepackage{booktabs}
\usepackage{multirow}
\usepackage{charter}
\usepackage{tikz}
\usepackage{caption}
\usepackage{makecell}
\usepackage{graphicx}
\usepackage{array} 
\usetikzlibrary{shapes.geometric, arrows, positioning}
\usepackage[style=mla,backend=biber]{biblatex} 
\definecolor{color1}{RGB}{0,0,90} 
\definecolor{color2}{RGB}{0,20,20} 

\usepackage{hyperref} 
\hypersetup{hidelinks,colorlinks,breaklinks=true,urlcolor=color2,citecolor=color1,linkcolor=color1,bookmarksopen=false,pdftitle={Title},pdfauthor={Author}}
\usepackage{cleveref}

\JournalInfo{\today} 
\Archive{Eric G} 

\PaperTitle{Prediction of Respiratory Syncytial Virus-Associated Hospitalizations Using Machine Learning Models Based on Environmental Data}

\Authors{Eric Guo \textsuperscript{1}*} 
\affiliation{\textsuperscript{1}\textit{Head-Royce School, Oakland, CA, United States}} 
\affiliation{*\textbf{Corresponding author}: \href{mailto:ericjiawenguo@gmail.com}{ericjiawenguo@gmail.com}} 

\newcommand{\keywordname}{Keywords} 

\Abstract{
Respiratory syncytial virus (RSV) is a leading cause of hospitalization among young children, with outbreaks strongly influenced by environmental conditions. This study developed a machine learning framework to predict RSV-associated hospitalizations in the United States (U.S.) by integrating wastewater surveillance, meteorological, and air quality data. The dataset combined weekly hospitalization rates, wastewater RSV levels, daily meteorological measurements, and air pollutant concentrations. Classification models, including CART, Random Forest, and Boosting, were trained to predict weekly RSV-associated hospitalization rates classified as \textit{Low risk}, \textit{Alert}, and \textit{Epidemic} levels. The wastewater RSV level was identified as the strongest predictor, followed by meteorological and air quality variables such as temperature, ozone levels, and specific humidity. Notably, the analysis also revealed significantly higher RSV-associated hospitalization rates among Native Americans and Alaska Natives. Further research is needed to better understand the drivers of RSV disparity in these communities to improve prevention strategies. Furthermore, states at high altitudes, characterized by lower surface pressure, showed consistently higher RSV-associated hospitalization rates. These findings highlight the value of combining environmental and community surveillance data to forecast RSV outbreaks, enabling more timely public health interventions and resource allocation. In order to provide accessibility and practical use of the models, we have developed an interactive R Shiny dashboard (\url{https://f6yxlu-eric-guo.shinyapps.io/rsv_app/}), which allows users to explore RSV-associated hospitalization risk levels across different states, visualize the impact of key predictors, and interactively generate RSV outbreak forecasts.
}

\renewcommand{\baselinestretch}{1.1}

\begin{document}

\flushbottom 
\maketitle  
\tableofcontents 
\thispagestyle{empty} 

\setlength{\parskip}{0.5em} 
\renewcommand{\baselinestretch}{1.2}

\section*{Introduction} 

\addcontentsline{toc}{section}{Introduction} 


Respiratory syncytial virus (RSV) is a respiratory virus that causes acute respiratory infections, typically with cold-like symptoms. However, it can also lead to severe illness such as bronchiolitis and respiratory tract failure. It is a leading cause of respiratory infections in infants and young children \autocite{oey2023}. Especially, children under five are the most vulnerable to severe outcomes. RSV poses a tremendous public health burden due to high rates of hospitalization and admission to intensive care units. In 2019, approximately 33 million RSV-related lower respiratory tract infections led to 3.6 million hospitalizations and 101,4000 deaths among children under five in the world \autocite{li2022}. In the U.S., the Centers for Disease Control and Prevention (CDC) estimates that RSV leads to 58{,}000 -- 80{,}000 hospitalizations annually among children younger than five \autocite{cdcsurveillance}. 

RSV transmits through airborne particles and via direct or indirect contact. In recent years, many studies have investigated the relationship between atmospheric conditions and RSV infections. For example, real-time weather data were used to predict RSV outbreaks in Salt Lake County, Utah. Temperature and wind speed were identified as the best predictors in a Naive Bayes (NB) model \autocite{walton2010}. Similarly, based on the study of the correlation between RSV bronchiolitis among children $\le 5$ years and climate conditions in Sousse, Tunisia, it was found that RSV infectivity was negatively correlated with temperature and humidity \autocite{brini2020}. More recently, multi-source data were used to predict pediatric RSV in the U.S., and precipitation and temperature were found to be correlated factors \autocite{yang2024}. In China, researchers developed a machine learning approach using environmental data to develop a nationwide respiratory virus infection risk prediction model, which showed the significant predictive effect of NO\textsubscript{2} levels and meteorological conditions \autocite{shi2025}. These studies have shown a strong relationship and predictive power of environmental factors for RSV outbreaks. However, these studies also have some limitations, such as the size of the datasets, restricted geographic coverage, and limited generalizability to other contexts.

In the paper, our goal was to develop a machine learning model to predict RSV outbreaks, specifically, RSV-associated hospitalizations in the U.S., based on real-time surveillance data from multiple national systems or agencies. Our proposed model will provide a thorough analysis of the disease nationwide, which will help our healthcare system to better prepare for future outbreaks, optimize the allocation of hospitalizations and intensive care resources, and save more lives with timely treatments.

\section{Methods}






\subsection{Data Sources}



\begin{figure*}[ht]
	\centering
	\includegraphics[width=1\linewidth]{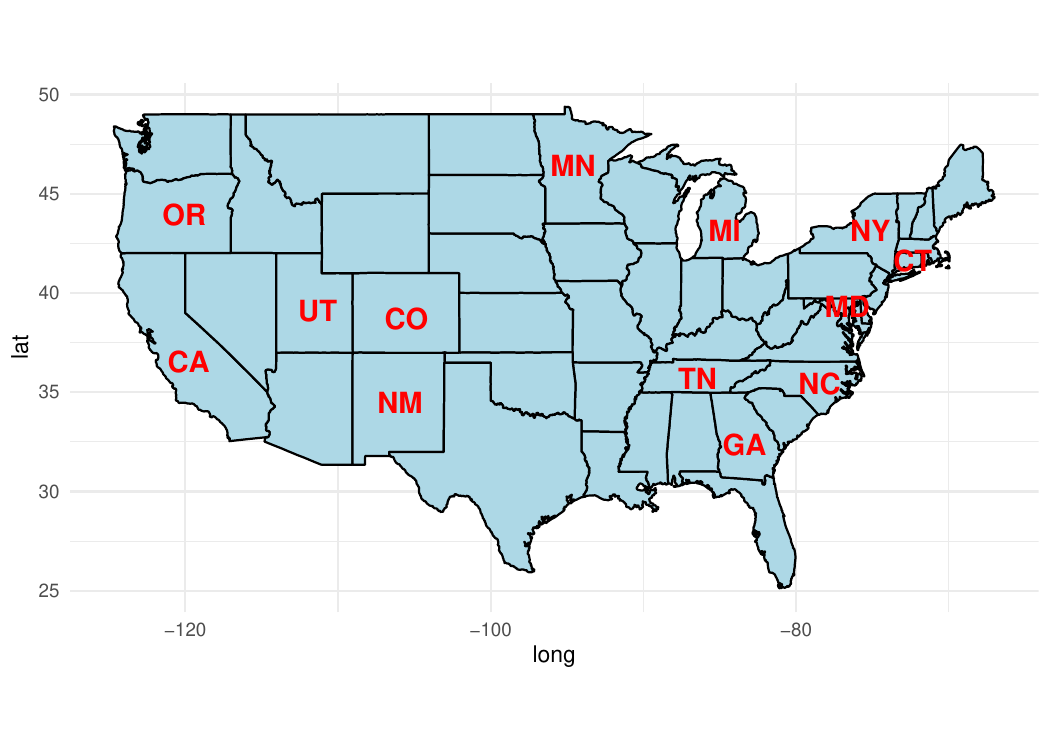} 
	\caption{RSV-NET covers 13 states in the U.S.}
	\label{fig:13states}
\end{figure*}

The weekly rates of laboratory-confirmed RSV-associated hospitalizations were collected by CDC RSV Hospitalization Surveillance Network \autocite{rsvnet}, which is a network that conducts active, population-based surveillance in children and adults. They cover 13 states across the U.S. (\autoref{fig:13states}). The weekly rates show how many people in the surveillance state are hospitalized due to RSV every week, compared to the total number of people residing in that area (Unit: number of hospitalizations/100,000 persons). For clarification, for children of a specific age group, the rate is reported per 100{,}000 children in that age group, not per 100{,}000 of the entire population. It is abbreviated as ``\verb|Response Rate|,'' ``\verb|Weekly Rate|," or ``\verb|Rate|" in the paper. 

Wastewater surveillance is a powerful system for monitoring infectious diseases within communities \autocite{kilaru2022}. Infected individuals can shed pathogens in their stool and urine, which end up in wastewater. In response to the COVID-19 pandemic, the CDC launched the National Wastewater Surveillance System \autocite{nwss} in 2020 to track the SARS-CoV-2 RNA levels in the U.S. wastewater. It later expanded the system to monitor other viral pathogens, including RSV. The wastewater viral activity level (WVAL) of RSV was monitored weekly, and the data are available from April 2022 to the present. Wastewater monitoring can detect RSV spreading within a community earlier than clinical testing, even before infected individuals seek medical attention. It can also detect infections within a community without symptoms. An increased WVAL indicates a higher risk of infection and is calculated by comparing the current amount of virus to a baseline and normalizing by the standard deviation. Thus, WVAL serves as a strong predictor of RSV outbreaks in a community. 

Daily meteorological data, such as temperature, humidity, precipitation, wind speed, surface pressure, and others, were obtained from the NASA \autocite{nasa}, a high-quality and freely accessible dataset from satellite observations and models.  

Daily outdoor air quality data were collected by the U.S. Environmental Protection Agency \autocite{epa}, including CO, NO\textsubscript{2}, Ozone, PM10, PM2.5, and SO\textsubscript{2} for each state.

Data from these reliable national surveillance systems were combined by matching date and location to create a high-quality dataset for developing our prediction model. The daily meteorological and outdoor air quality measurements were averaged by week to align with weekly RSV-associated hospitalization rate and WVAL. Some of the key variables in the combined dataset are summarized in \autoref{tab:keyvariables}, with their definitions and units.



\begin{table}[!htb]
	\begin{center}
	\begin{footnotesize}
		\caption{Summary of key variables}
		\begin{tabular}{|p{2cm}|>{\raggedright\arraybackslash}p{4cm}|>{\raggedright\arraybackslash}p{1cm}|}
			\hline
			\textbf{Variable} & \multicolumn{1}{c|}{\textbf{Definition}} & \multicolumn{1}{c|}{\textbf{Unit}} \\
			\hline
			 Rate & Weekly rates of laboratory-confirmed RSV-associated hospitalizations & /100,000 persons \\
			\hline
			WVAL* & Wastewater RSV level & — \\
			\hline
			PRECTOTCORR* & Daily precipitation & mm/day \\
			PS* & Surface pressure & kPa \\
			QV2M* & Specific humidity at 2 meters above the ground & g/kg \\
			RH2M* & Relative humidity at 2 meters above the ground (percentage of moisture in the air relative to maximum possible at that temperature) & \% \\
			T2M* & Air temperature at 2 meters above the ground & °C \\
			T2MDEW & Dew/frost point at 2 meters above the ground (temperature at which air becomes saturated and water condenses or freezes) & °C \\
			T2MWET & Wet-bulb temperature at 2 meters above the ground (lowest temperature achievable by evaporative cooling) & °C \\
			TS & Earth skin temperature (radiative temperature of the uppermost layer of the Earth's surface) & °C \\
			WD10M* & Wind direction at 10 meters above the ground & degrees \\
			WS10M* & Wind speed at 10 meters above the ground & m/s \\
			WS2M & Wind speed at 2 meters above the ground & m/s \\
			\hline
			CO* & Carbon monoxide concentration & ppm \\
			NO\textsubscript{2}* & Nitrogen dioxide concentration & ppm \\
			Ozone* & Ozone concentration & ppm \\
			PM10* & Particulate matter $\leq 10~\mu\mathrm{m}$ & $\mu\mathrm{g/m^3}$ \\
			PM2.5* & Particulate matter $\leq 2.5~\mu\mathrm{m}$  & $\mu\mathrm{g/m^3}$ \\
			SO\textsubscript{2}* & Sulfur dioxide concentration & ppb \\
			\hline
			RSV Season* &  Yes (Nov - April); No (May - Oct) & \\
			\hline
		\end{tabular}
		\caption*{\footnotesize Note: * indicates the predictors to be included for model development.}
		\label{tab:keyvariables}
	\end{footnotesize}
	\end{center}
\end{table}

\subsection{Analysis Methods}
A machine learning framework was developed to predict national RSV-associated hospitalizations, enabling real-time and precise assessment of hospitalization trends and potential stress on the healthcare system using national surveillance databases. Multiple machine learning approaches were evaluated to identify the best-performing model for prediction. By leveraging publicly available wastewater, meteorological, and air quality data, a comprehensive model was constructed to support public health planning and decision-making. Data cleaning and analysis were mostly conducted in RStudio. Negative values of CO, NO\textsubscript{2} and SO\textsubscript{2} due to instrument measurement errors were imputed as zero. The flowchart of the model-building process is shown in \autoref{fig:ml_rsv}. 

\tikzstyle{input} = [rectangle, rounded corners, minimum width=3cm, minimum height=1cm,text centered, draw=black, fill=orange!40]
\tikzstyle{process} = [rectangle, minimum width=3cm, minimum height=2cm, text centered, draw=black, fill=blue!10]
\tikzstyle{arrow} = [thick,->,>=stealth]
\tikzstyle{prob} = [rectangle, minimum width=2.5cm, minimum height=0.8cm, text centered, draw=black, fill=white!90]
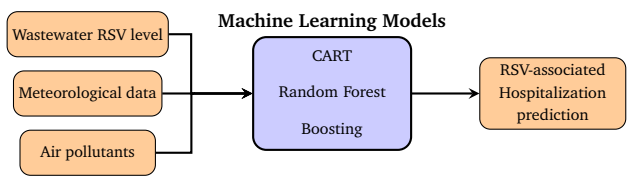
\begin{figure}[h]
	\centering
\begin{tikzpicture}[node distance=0.3cm, scale=0.6, transform shape]
	\node (patient) [input] {Wastewater RSV level};
	\node (meteo) [input, below=0.3cm of patient] {Meteorological data};
	\node (pollut) [input, below=0.3cm of meteo] {Air pollutants};
	\node (mlbox) [draw, thick, rectangle, rounded corners=5pt, right=2cm of meteo, 
	minimum width=3.5cm, minimum height=2.5cm, 
	align=center, text width=3cm, label=above:{\textbf{\large Machine Learning Models}},
	fill=blue!20] 
	{CART\\[0.3cm]Random Forest\\[0.3cm]Boosting};
	\draw [arrow] (patient.east) -- ++(0.5,0) |- (mlbox.west);
	\draw [arrow] (meteo.east) -- ++(0.7,0) |- (mlbox.west);
	\draw [arrow] (pollut.east) -- ++(0.79,0) |- (mlbox.west);
\node (output) [input, right=1.5cm of mlbox, text width=3cm, align=center] 
{RSV-associated Hospitalization prediction};
	\draw [arrow] (mlbox.east) -- ++(0.7,0) -- (output.west);
\end{tikzpicture}
\vspace{5mm} 
\caption{Flowchart of model-building process}
\label{fig:ml_rsv}
\end{figure}



\section{Results}


\subsection{Data Characteristics}
The overall weekly rate across 13 states was visualized over time by demographic groups, including gender, age, and race/ethnicity groups (\autoref{tab:variables}). 
\begin{table}[!htb]
	\begin{small}
	\caption{Summary of demographic groups}
	\begin{center}
	\begin{tabular}{|l|p{4.5cm}|}  
		\hline
		\textbf{Variable} & \textbf{Groups} \\
		\hline
		Gender & Male, Female \\
		\hline
		Age group 1 (years) & 0--4, 5--17, 18--49, 50--64, $\geq$65 \\
		Age group 2 (years) & 0--17, $\geq$18 \\
		\hline
		Race/Ethnicity & White, non-Hispanic\\
		& AI/AN, non-Hispanic\\
		& Black, non-Hispanic\\
		& A/PI, non-Hispanic\\
		& Hispanic \\
		\hline
	\end{tabular}
	\end{center}
	\label{tab:variables}
	\end{small}
\end{table}

\begin{figure}[!htb]
	\centering
	\includegraphics[width=1\linewidth]{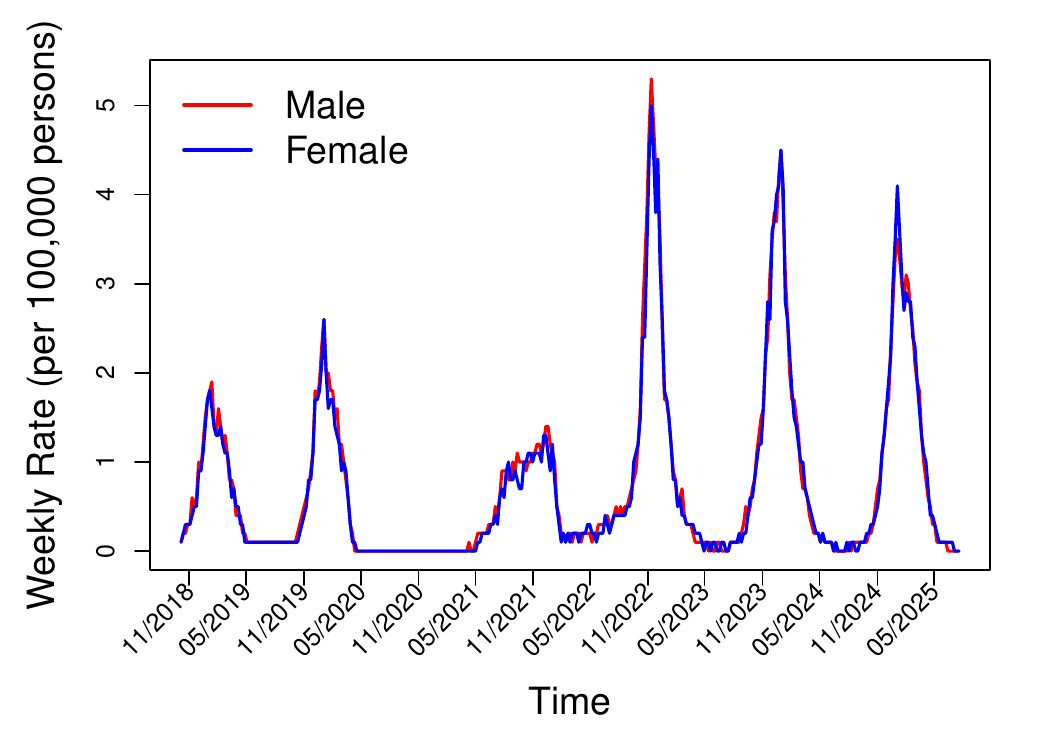} 
	\caption{Weekly rate by gender}
	\label{fig:timetrendbygender}
\end{figure}

In \autoref{fig:timetrendbygender}, the time trend showed the seasonal pattern of RSV-associated hospitalizations, which typically occurred between November and April, and peaked during the winter season each year. However, an unusual pattern was observed between November 2020 and April 2021, during which an RSV epidemic did not occur. Instead, the expected seasonal outbreak was delayed and overlapped with the subsequent RSV season from November 2021 to April 2022. This deviation was likely due to the first recommendation of CDC for public mask use in April 2020 during the COVID-19 pandemic, which also prevented the transmission of other airborne infectious diseases, including RSV. Furthermore, the seasonal trend was similar for both males and females. 

\subsubsection{High Response Rate for Children (0--4 years)}
\begin{figure*}[!htb]
	\centering
	\includegraphics[width=1\linewidth]{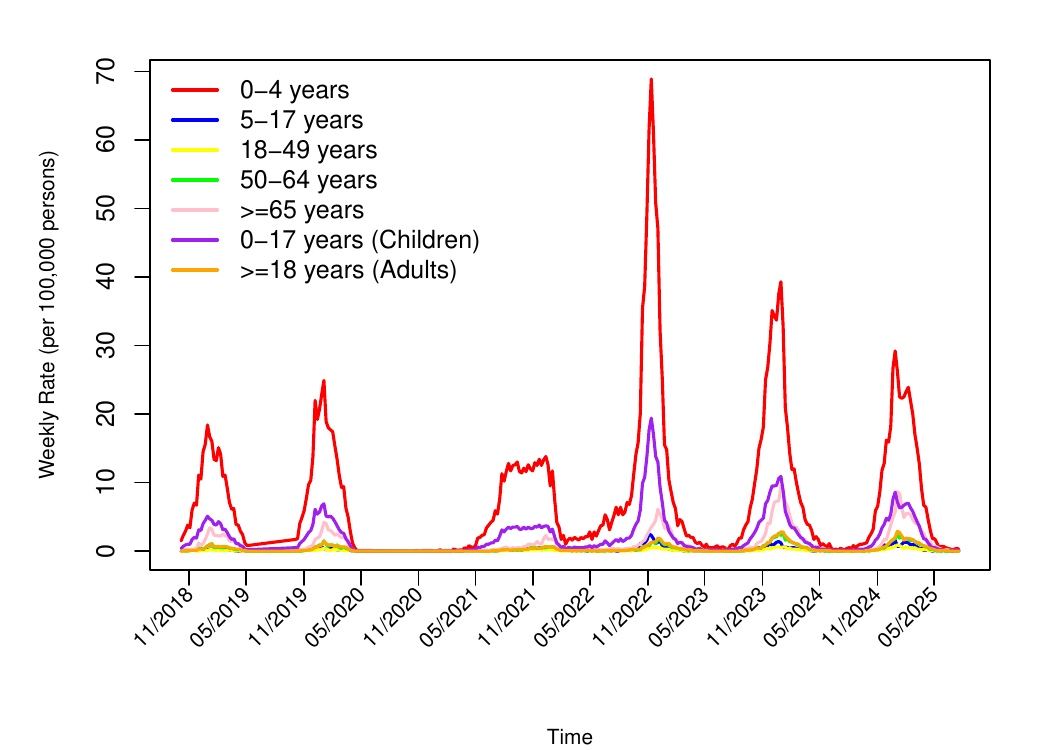} 
	\caption{Weekly rate by age groups}
	\label{fig:timetrendbyage}
\end{figure*}

However, significant differences existed among age and race/ethnicity groups. Among all age groups, children (0--4 years) experienced the highest rate, which indicated that this group was most impacted by RSV (\autoref{fig:timetrendbyage}). The age group of 0--17 years experienced the second-highest rate. Since the pediatric population (0--4 years) was most impacted, the following analysis and model development focused only on this group. 

\subsubsection{High Response Rate for American Indians and Alaska Natives}
\begin{figure*}[!htb]
	\centering
	\includegraphics[width=1\linewidth]{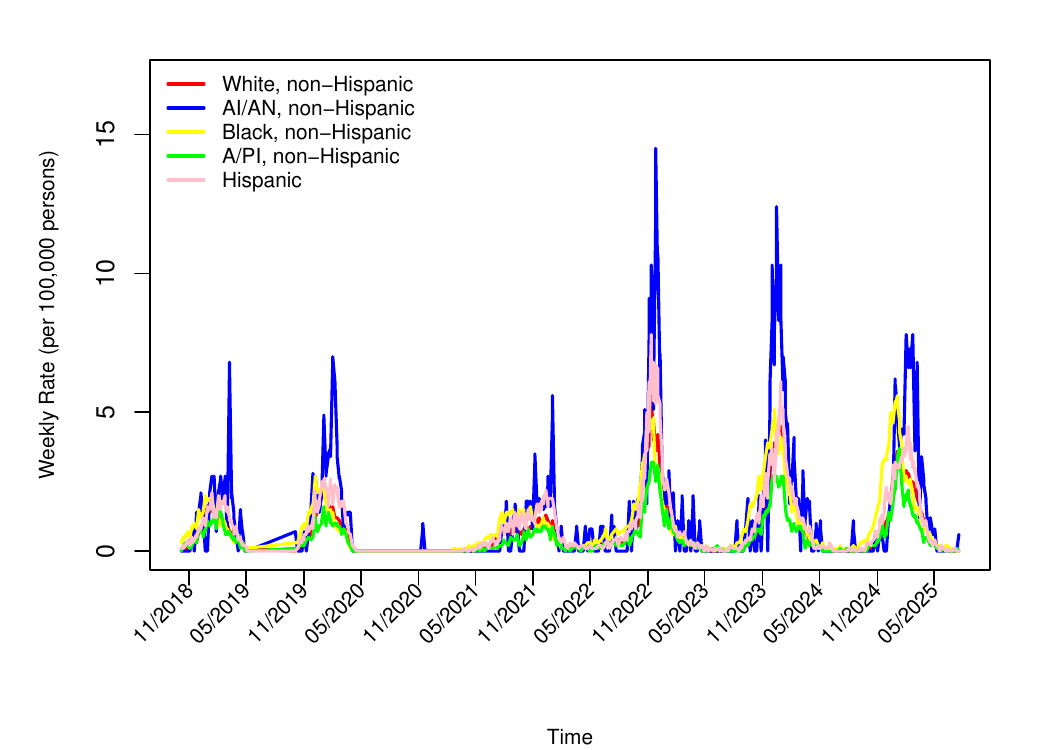} 
	\caption{Weekly rate by race/ethnicity groups}
	\label{fig:timetrendbyrace}
\end{figure*}

Among racial and ethic groups, the American Indians and Alaska Natives (AI/AN, non-Hispanic) had substantially higher rates than other groups (\autoref{fig:timetrendbyrace}). This finding has also been reported by other researchers, who found that household crowding was associated with an increased risk of hospitalization \autocite{bulkow2002rsv}. In general, Indigenous populations are more likely to be displaced from their ancestral lands to face poverty, poor living conditions, lower education levels, and higher unemployment rates \autocite{chang2015lung}. These socioeconomic factors contribute to higher rate of lower respiratory infections, with housing conditions emerging as a particularly important reason. A disproportionately high percentage of Indigenous families live in overcrowded homes with poor ventilation, which not only facilitates the spread of respiratory infections but also increases exposure to tobacco smoke and other infectious cofactors such as secondary bacterial infections accompanying viral illnesses \autocite{basnayake2017global}. Additional risk factors, including malnutrition, limited access to clean water, and poor sanitation, further jeopardize overall health and increase infection risk \autocite{basnayake2017global}. These socioeconomic and environmental disparities raise serious concerns about the health of Indigenous communities, and further research should focus on states with large Indian reservations to better understand and address these inequities.

\subsubsection{Data Distribution}
\begin{figure}[!htb]
	\centering
	\vspace{1.5cm}
	\includegraphics[width=1\linewidth]{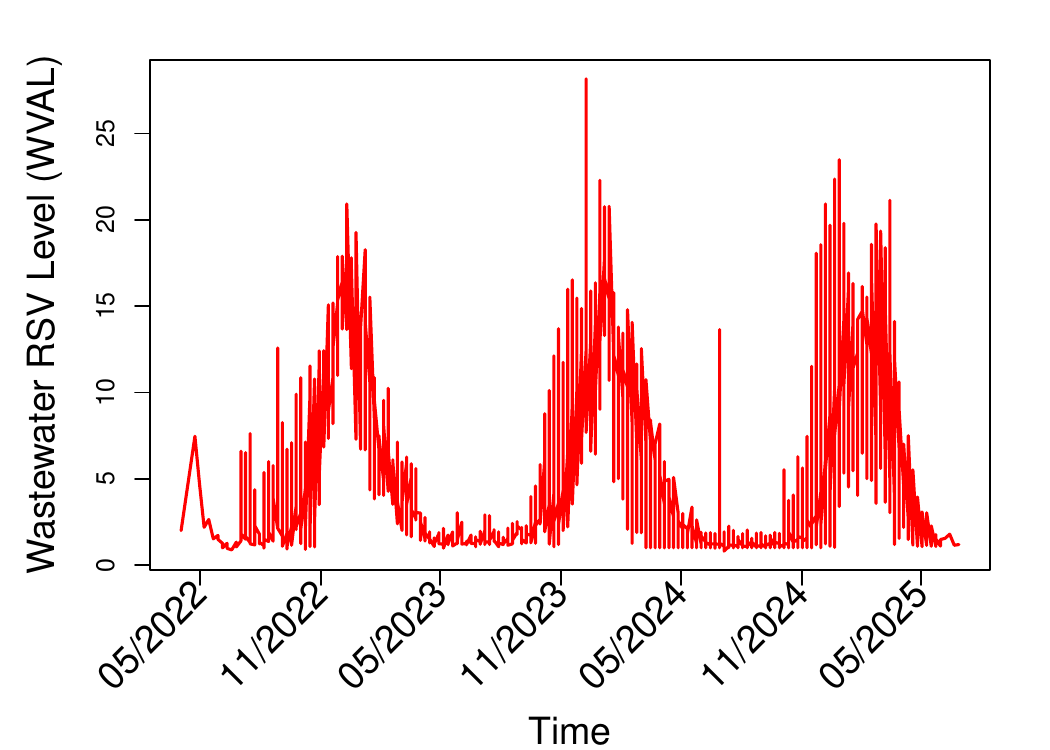} 
	\vspace{0.05cm}
	\caption{Weekly wastewater RSV levels (WVAL)}
	\label{fig:timetrendwastewater}
\end{figure}

We also conducted an exploratory analysis of other key variables. RSV \verb|WVAL| showed a clear seasonal pattern that closely aligned with the seasonal trend of weekly RSV-associated hospitalization rates, suggesting that \verb|WVAL| could serve as a strong predictor of the weekly rate (\autoref{fig:timetrendwastewater}).

\begin{figure}[ht]
	\centering
	\vspace{0.8cm}
	\includegraphics[width=1\linewidth]{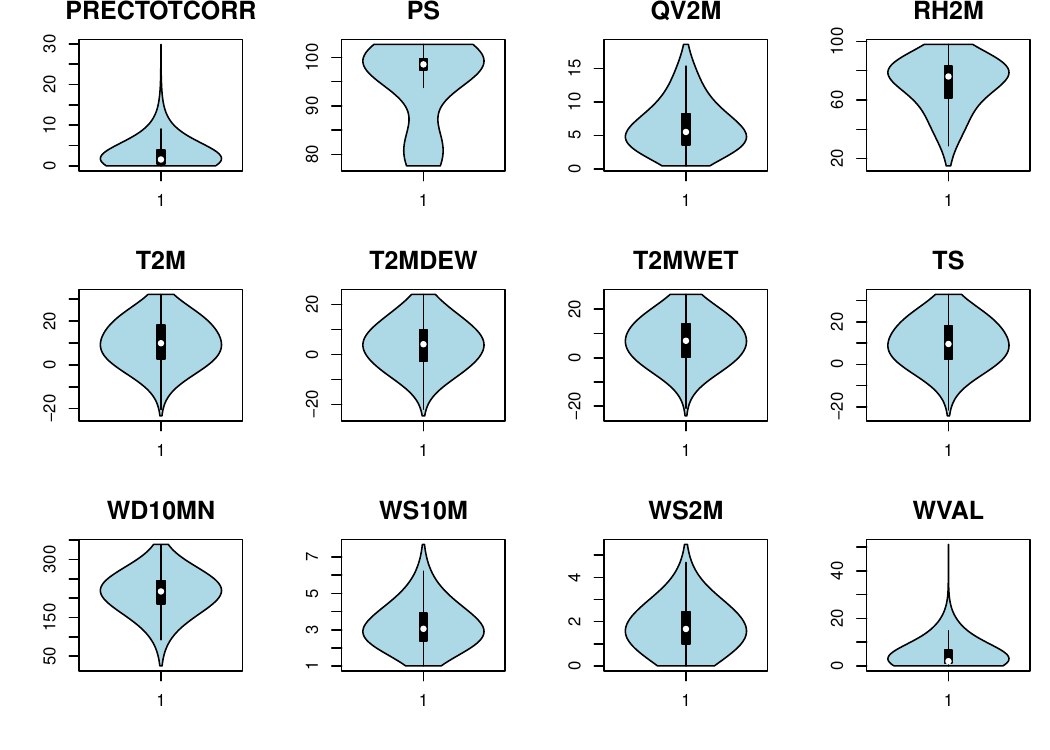} 
	\caption{Distribution of meteorological variables and WVAL.}
	\label{fig:meteorologicalvariables}
\end{figure}

\begin{figure}[ht]
	\centering
	\includegraphics[width=1\linewidth]{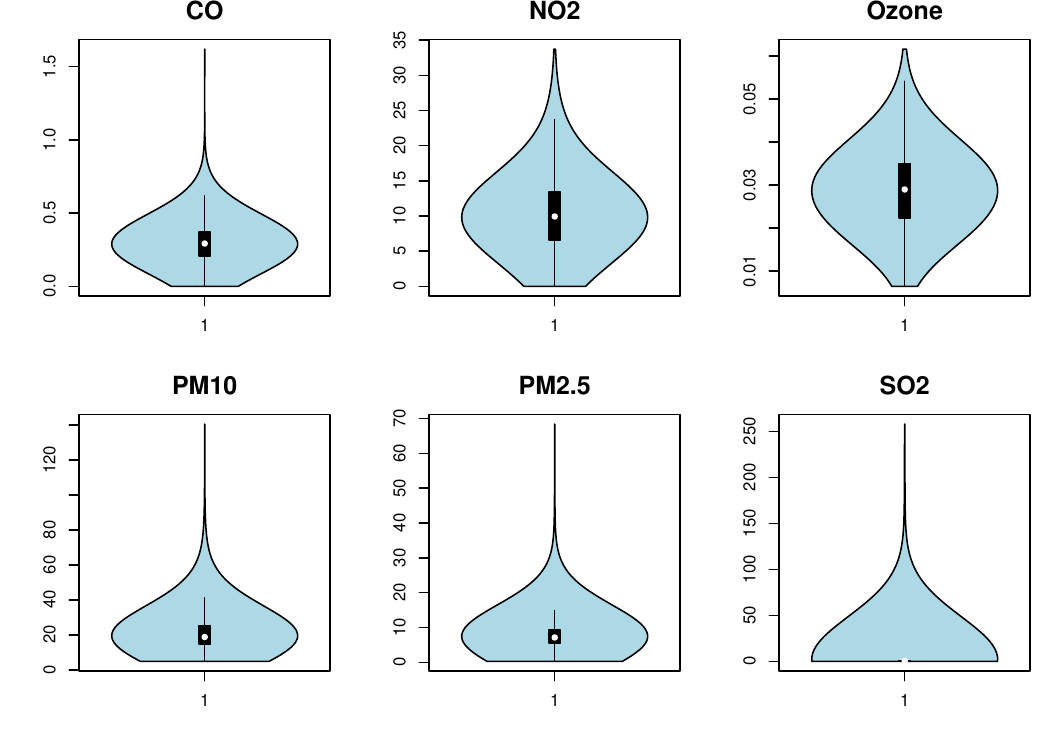} 
	\caption{Distribution of air pollutant variables}
	\label{fig:pollutionvariables}
\end{figure}

In addition, we investigated the distribution of \verb|WVAL|, meteorological variables, and air pollutant levels using violin plots to visualize the spread and shape of the data (\autoref{fig:meteorologicalvariables} and \autoref{fig:pollutionvariables}). Examining these distributions allowed us to understand the shape, spread, and skewness of the data, identify patterns and anomalies, and choose appropriate methods for subsequent analyses. The distributions of the temperature-related variables \texttt{T2M}, \texttt{T2MDEW}, \texttt{T2MWET}, and \verb|TS| were similar to each other. Likewise, the wind speed variables (\verb|WS10M| and \verb|WS2M|) showed comparable patterns. The \verb|WVAL| data were highly skewed with many observations around zero. In addition, \verb|PS| displayed a bimodal distribution with two separate ranges of values.

%
%
%

\subsection{Response and Predictors}
\begin{figure*}[!htb]
	\centering
	\includegraphics[width=1\linewidth]{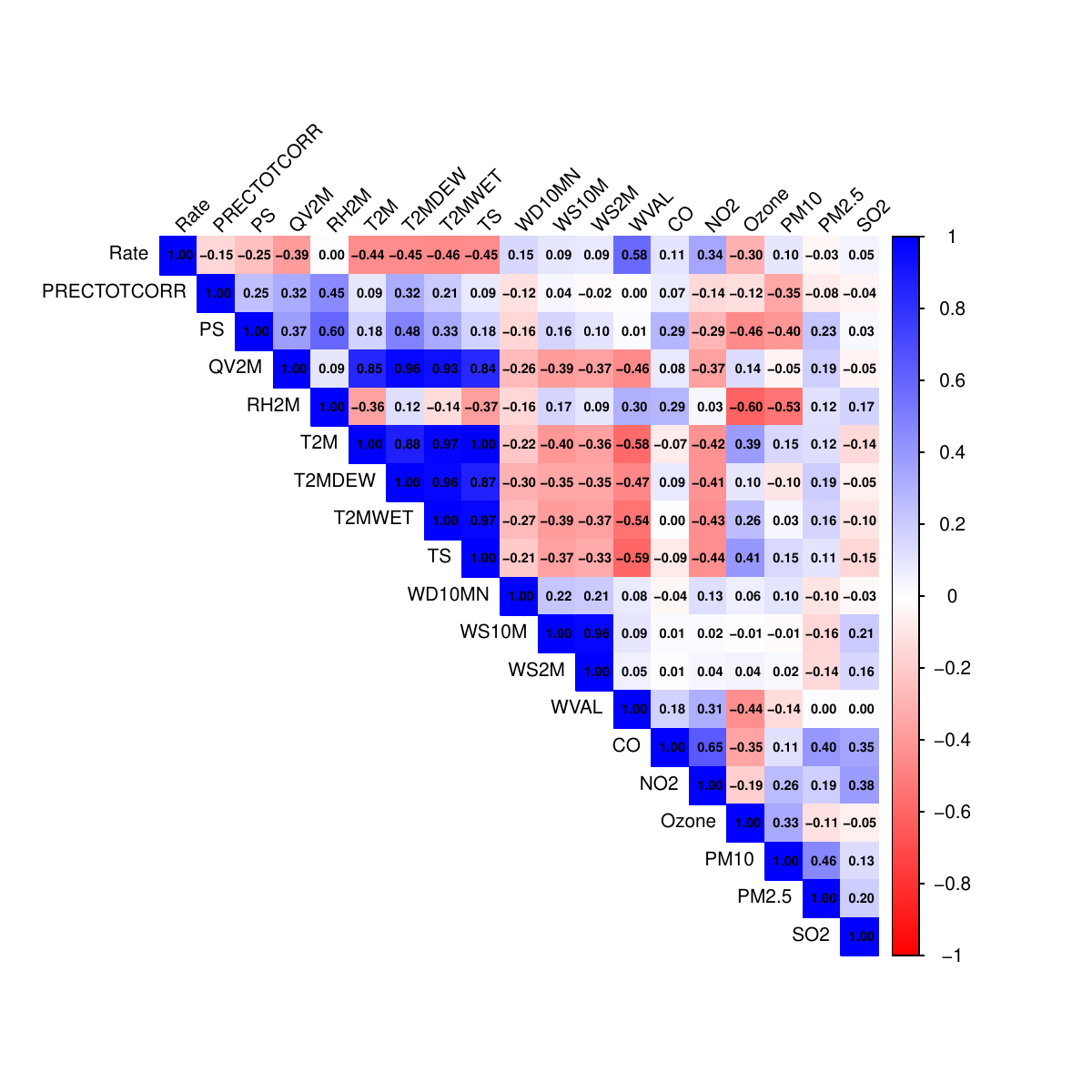} 
	\caption{Correlation matrix}
	\label{fig:correlationmatrix}
\end{figure*}

We investigated the correlations among all variables to identify predictors that were highly correlated with the response variable. Additionally, we assessed correlations among the predictors to avoid including highly collinear predictors in the model. The resulting correlation matrix showed several interesting observations (\autoref{fig:correlationmatrix}). 

The response variable \verb|Rate| was negatively correlated with temperature-related variables (\texttt{T2M}, \texttt{T2MDEW}, \texttt{T2MWET}, and \texttt{TS}), with correlation coefficients ranging from -0.44 to -0.46 (\autoref{fig:correlationmatrix}), consistent with RSV-associated hospitalizations peaking in fall and winter (November to April; Figures  ~\ref{fig:timetrendbygender} -- \ref{fig:timetrendbyrace}). It was also negatively correlated with specific humidity (\verb|QV2M|; correlation coefficient = -0.39), which measures the mass of water vapor per unit mass of air (g/kg), suggesting that RSV transmits more readily under dry air conditions with low humidity. Negative correlations were also observed with \verb|Ozone| (-0.30) and \verb|PS| (-0.25). On the contrary, the response was positively correlated with \verb|WVAL| (0.58) and \verb|NO|\textsubscript{2} (0.34). 

%
Among all predictors, the four temperature variables were highly associated, with correlation coefficients ranging from 0.87 to 1.00. Wind speeds at 2 meters and 10 meters were also highly correlated (0.96). For highly correlated predictors, only one of them was selected for inclusion in the model to avoid multicollinearity. Since \verb|T2M| (air temperature at 2 meters above the ground) is a standard and widely used measure of ambient temperature, it was included in the model. For wind speed, \verb|WD10M| was included in the model since it is more representative of the general environmental condition, whereas \verb|WS2M| is closer to the ground and more affected by obstacles such as trees and buildings. The predictors selected for model development are indicated in \autoref{tab:keyvariables}.

\begin{figure}[!htb]
	\centering
	\includegraphics[width=1\linewidth]{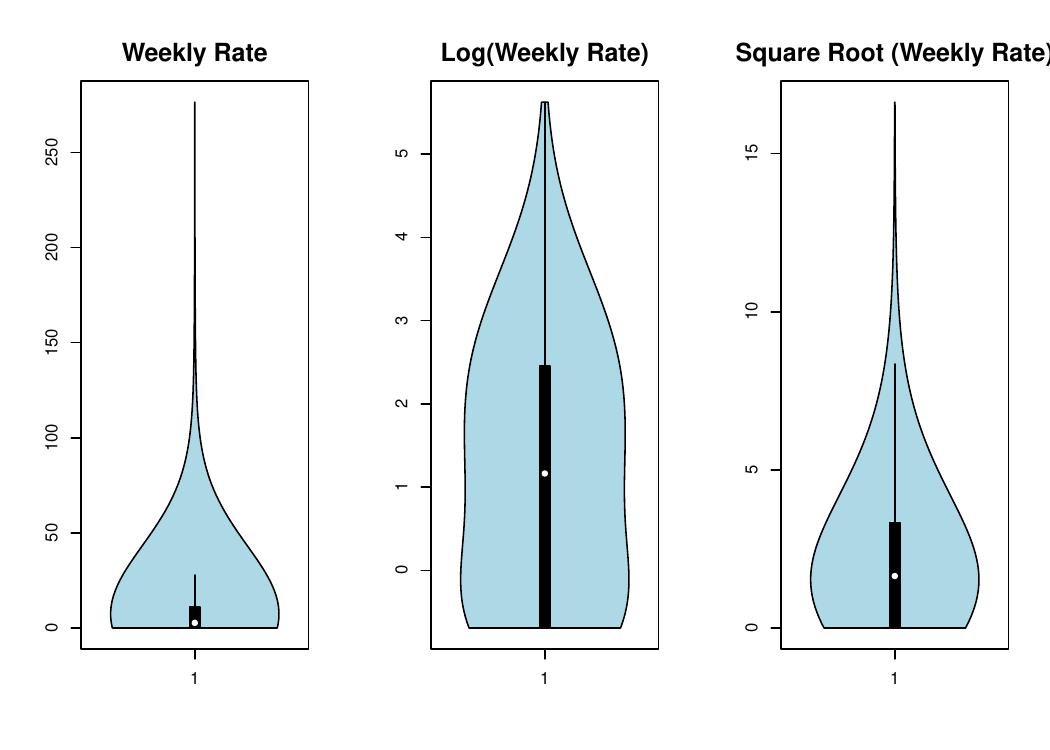} 
	\caption{Distribution of response variable}
	\label{fig:responsedistribution}
\end{figure}

The response rate contained many zero values, resulting in a highly skewed distribution. We explored different transformations, including logarithmic and square root. However, neither of them was able to handle the heavy tail of zeros well (\autoref{fig:responsedistribution}). 

In order to address this issue, we converted the continuous response rate into a class variable by using thresholds to categorize it into three classes (\autoref{tab:cutoffs}).

\begin{table}[!htb]
	\centering
	\begin{small}
		\caption{Categories of class response variable}
		\begin{tabular}{|l|p{5cm}|}
			\hline
			\textbf{Class} & \textbf{Thresholds (per 100,000 persons)} \\
			\hline
			Low risk & 0--5 \\
			Alert & $>$ 5 and $<$ 20 \\
			Epidemic & $\geq 20$ \\
			\hline
		\end{tabular}
		\label{tab:cutoffs}
	\end{small}
\end{table}


\subsection{Model Performance and Comparison}

We applied three machine learning approaches, Classification and Regression Trees (CART), Random Forest, and Boosting, to model the response. The data were split into training (80\%) and testing sets (20\%), with the split preserving class proportions in both subsets. The summary statistics of the response and predictors in the two sets were comparable (\autoref{tab:comp_summary}).

\begin{table}[!htb]
	\centering
	\begin{footnotesize}
	\caption{Summary statistics of training and testing sets}
	\begin{tabular}{|p{2cm}|c|c|c|c|}
		\hline
		\multirow{2}{*}{\textbf{Variables}} & \multicolumn{2}{c|}{\textbf{Training set}} & \multicolumn{2}{c|}{\textbf{Testing set}} \\
		\cline{2-5}
		& \textbf{Category} & \textbf{Percent} & \textbf{Category} & \textbf{Percent} \\
		\hline
		Rate & Low risk & 59.7\% & Low risk & 60.0\%\\
		& Alert & 19.3\% & Alert & 19.1\%\\
		& Epidemic & 20.9\% & Epidemic & 21.0\%\\
		\hline
		RSV Season & Yes & 52.1\% & Yes & 51.0\%\\
		 & No & 47.9 \% & No & 49.1\%\\
		\hline
		& \textbf{Mean} & \textbf{SD} & \textbf{Mean} & \textbf{SD} \\
		\hline
		WVAL & 4.96 & 4.87 & 5.35 & 5.45 \\
		\hline
		PRECTOTCORR & 2.16 & 2.45 & 2.34 & 2.63\\
		PS &92.72& 8.92& 92.25& 8.86\\
		QV2M & 6.63& 3.75&6.33 & 3.67\\
		RH2M & 65.79& 16.18& 67.03& 16.00\\
		T2M & 11.83 & 10.08 & 10.46 & 10.52 \\
		WD10MN & 210.78 &45.60 & 213.54 & 47.72 \\
		WS10M &3.51 & 1.12& 3.57& 1.14\\
		\hline
		CO & 0.32& 0.13& 0.32& 0.12\\
		NO\textsubscript{2} & 10.30& 5.13& 9.94& 5.00\\
		Ozone & 0.03& 0.01& 0.03& 0.01\\
		PM10 & 21.88& 9.93& 20.87& 9.45\\
		PM2.5 & 8.09&4.18 & 7.80& 3.82\\
		SO\textsubscript{2} & 0.60& 0.40& 0.64& 0.48\\		
		\hline
	\end{tabular}
	\label{tab:comp_summary}
	\end{footnotesize}

\end{table}
We used a 10-fold cross-validation to train the models on the training set. The selected model was then applied to the testing set to predict both the class and its probability. Prediction performance was evaluated based on the confusion matrix, F-1 score, and Receiver Operating Characteristic (ROC) curves. Using the same training and testing sets, we compared the three approaches to determine which method performed best for our application. 

F-1 score is defined as the harmonic mean of precision and recall \autocite{sasaki2007}. 

\begin{equation}
	\text{F-1 Score} = 2 \cdot \frac{\text{Precision} \cdot \text{Recall}}{\text{Precision} + \text{Recall}}
\end{equation}

\begin{equation}
	\text{Precision} = \frac{\text{TP}}{\text{TP} + \text{FP}}
\end{equation}

\begin{equation}
	\text{Recall} = \frac{\text{TP}}{\text{TP} + \text{FN}}
\end{equation}

where TP = True Positives, FN = False Negatives, and FP = False Positives. The value of F-1 score lies between 0 and 1, and a higher value means a better balance between precision and recall.

\subsubsection{CART}
\begin{figure*}[!htb]
 	\centering
 	\includegraphics[width=1\linewidth]{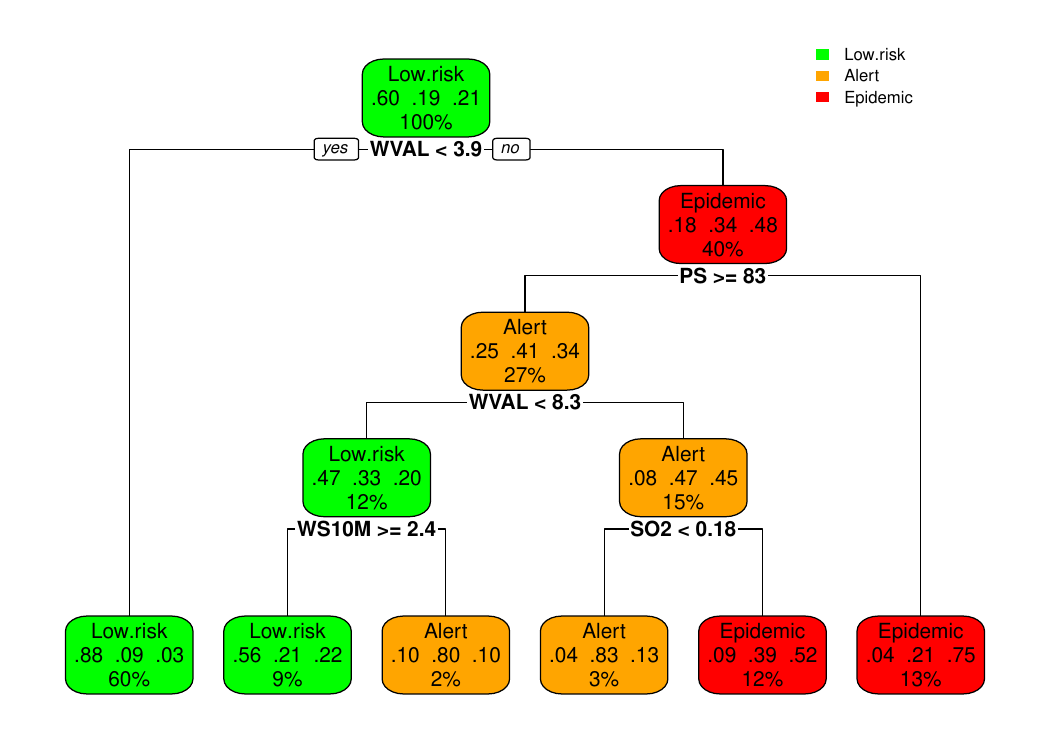} 
 	\caption{Selected CART tree}
 	\label{fig:CARTfinal}
 \end{figure*}

CART, a decision tree algorithm for predictive modeling, was used here to model the categorical response variable. A classification tree was constructed to predict the response classes (\textit{Low risk, Alert} and \textit{Epidemic}). The selected tree through a 10-fold cross-validation is shown in \autoref{fig:CARTfinal}.
 
\begin{figure}[!htb]
 	\centering
 	\includegraphics[width=1\linewidth]{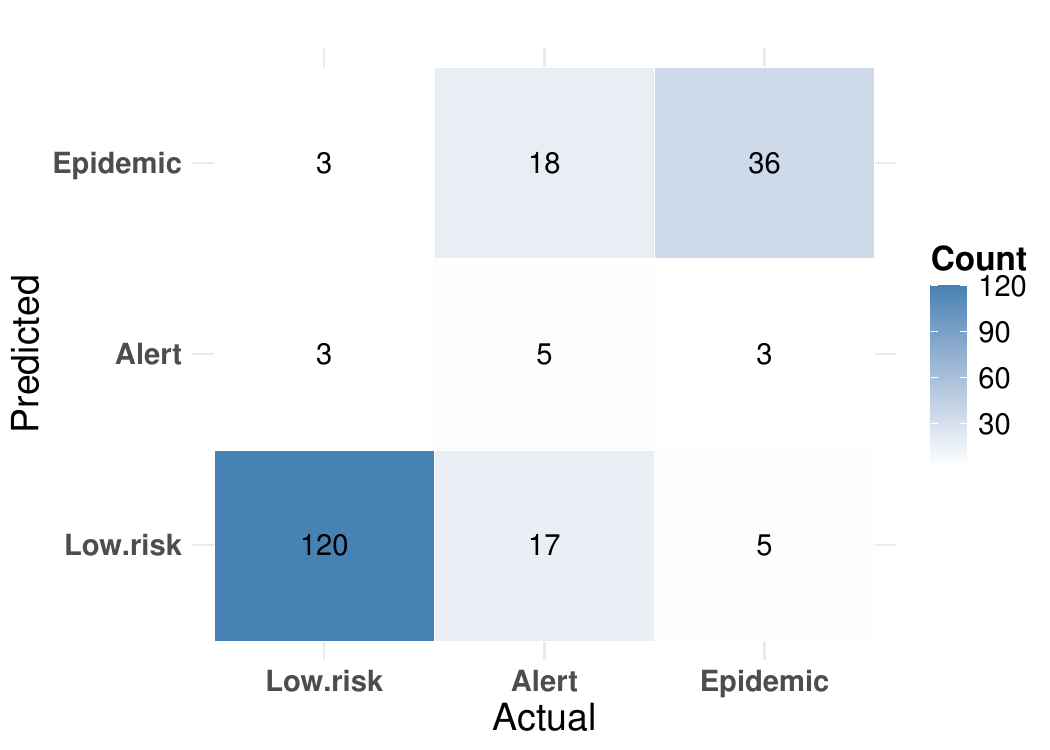} 
 	\caption{CART confusion matrix}
 	\label{fig:confusionmatrix}
 \end{figure}

 The first split was based on \verb|WVAL|, with observations below 3.9 being classified as \textit{Low risk}, while observations with \verb|WVAL|  $\geq$  3.9 progressed to further splits. Among observations with elevated \verb|WVAL|, the next split was based on \verb|PS|, where observations below 83 were classified as \textit{Epidemic}, whereas the observations with PS $\ge$ 83 required additional splitting based on \verb|WVAL|, \verb|WS10M|, and \verb|SO2|. Overall, the results showed that \verb|WVAL| was the strongest predictor, with meteorological and air quality data further refining the prediction. The CART model selected achieved an overall prediction accuracy of 0.767 on the testing set. The confusion matrix is shown in \autoref{fig:confusionmatrix}.
 
\begin{figure}[!htb]
 	\centering
 	\includegraphics[width=1\linewidth]{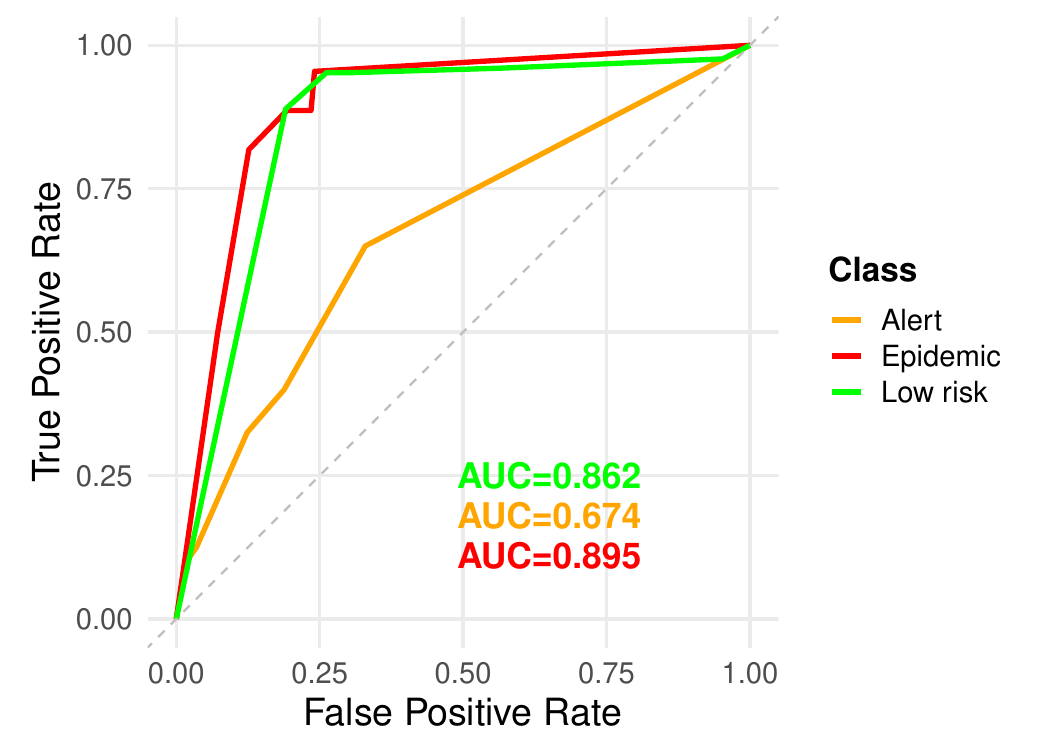} 
 	\caption{CART ROC}
 	\label{fig:cartROC}
 \end{figure}

 The ROC Curves ({\autoref{fig:cartROC}}) showed \textit{True Positive Rate} versus \textit{False Positive Rate} for each class, and the area under the curve (AUC) is 0.862, 0.674, and 0.895 for \textit{Low risk}, \textit{Alert}, and \textit{Epidemic}, respectively, which indicates better performance for the \textit{Low risk} and \textit{Epidemic} classes than \textit{Alert}. Overall, these results suggested strong discrimination between the classes. 
 
 
The variable importance analysis confirmed that \verb|WVAL| contributed the most to model performance, followed by \texttt{QV2M, T2M}, and \verb|RSV Season| (\autoref{tab:model_performance_classes}). These findings suggest that the wastewater surveillance, combined with meteorological and air quality data, can provide early warnings for the risk of RSV-associated hospitalizations.

\newcolumntype{P}[1]{>{\raggedright\arraybackslash}p{#1}}
\begin{table*}[!htb]
	\begin{small}
		\begin{center}
		\caption{Model performance comparison}
		\begin{tabular}{|p{4cm}|c|c|c|}
			\hline
			\textbf{Metric} & \textbf{CART} & \textbf{Random Forest} & \textbf{Boosting} \\
			\hline
			Accuracy   & 0.767 & 0.810 & 0.790 \\
			\hline
			F-1 Score   & 0.724 & 0.797 & 0.783 \\
			\hline
			AUC & & & \\
			\hspace{0.5cm}   Low           & 0.862 & 0.962 & 0.949 \\
			\hspace{0.5cm}   Alert          & 0.674 & 0.859 & 0.845\\
			\hspace{0.5cm}   Epidemic       & 0.895 & 0.966 & 0.953 \\
			\hline
			\makecell[l]{Top 4 important variables \\ \\ \\ \\} & 
			\makecell[c]{WVAL \\ QV2M \\ T2M \\ RSV Season} &
			\makecell[c]{WVAL \\ T2M \\ Ozone \\ QV2M} &
			\makecell[c]{WVAL \\ Ozone \\ PS \\ QV2M} \\
			\hline
		\end{tabular}
		\label{tab:model_performance_classes}
		\end{center}
	\end{small}
\end{table*}

\subsubsection{Effect of Surface Pressure on Response}
\begin{figure}[!htb]
	\begin{center}
	\includegraphics[width=1\linewidth]{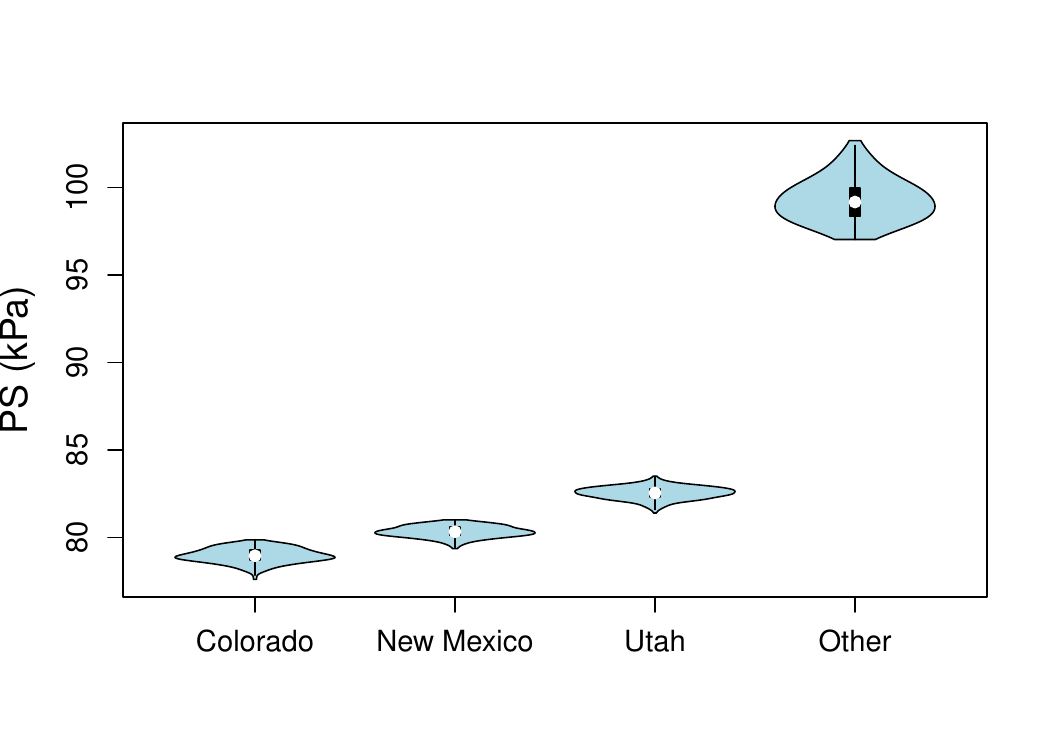} 
	\caption{PS for high-altitude states}
	\label{fig:psdistributionforhighaltitude}
	\end{center}
\end{figure}

\begin{figure}[!htb]
	\begin{center}
	\includegraphics[width=1\linewidth]{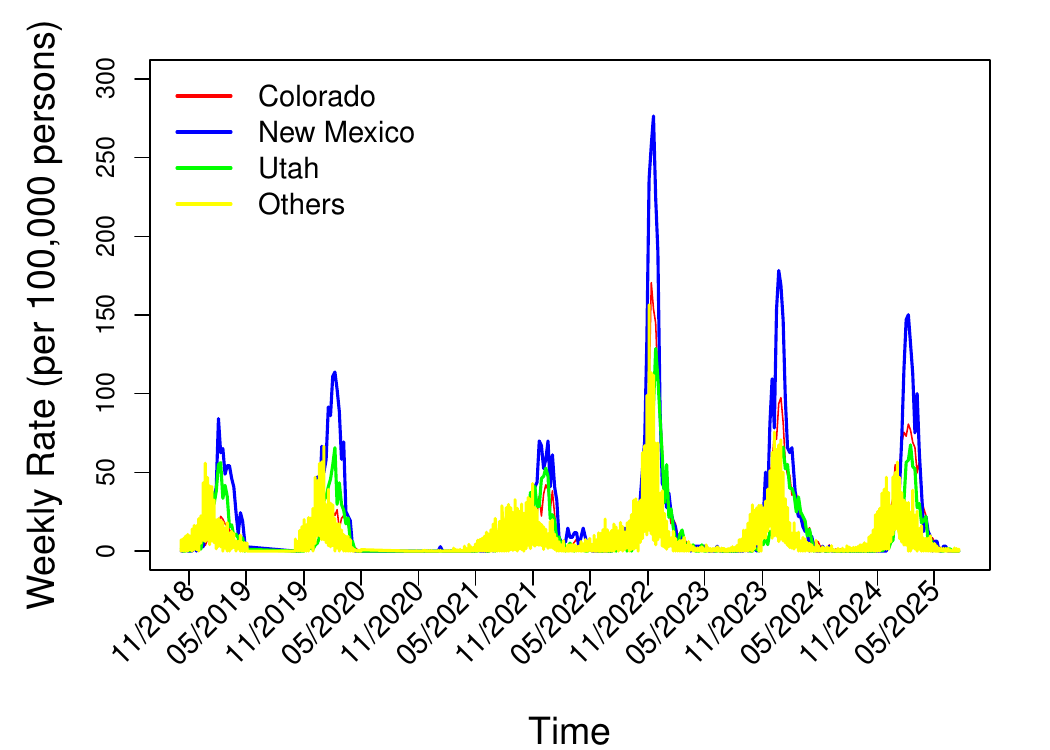} 
	\caption{Weekly rate for high-altitude states}
	\label{fig:pstimetrendforhighaltitude}
	\end{center}
\end{figure}

\begin{figure*}[!htb]
	\begin{center}
	\includegraphics[width=1\linewidth]{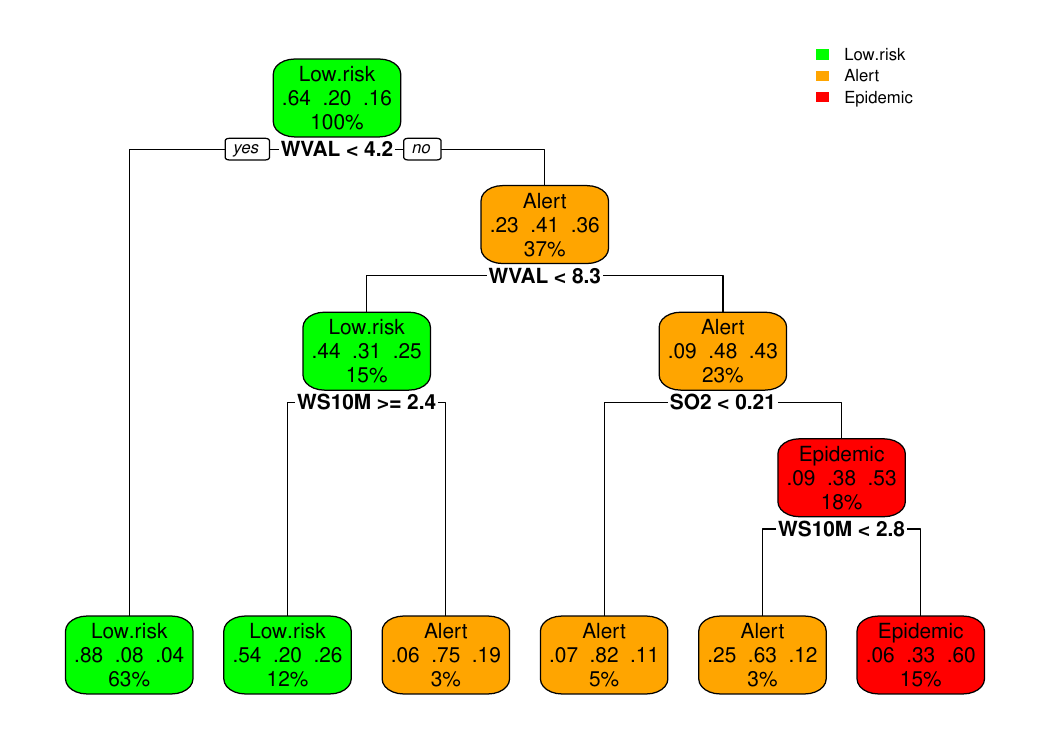} 
	\caption{Selected CART tree with the three states removed}
	\label{fig:finalcartmodel}
	\end{center}
\end{figure*}

It was found that all cases with surface pressure \verb|PS| $< 83$ corresponded to the high-altitude states of Colorado, New Mexico, and Utah (\autoref{fig:psdistributionforhighaltitude}), which also showed higher hospitalization rates than the other states (\autoref{fig:pstimetrendforhighaltitude}). This finding explained the bimodal distribution of \verb|PS| observed earlier and why these observations were classified as \textit{Epidemic}. In the model, \verb|PS| served as a geographic indicator to represent regional variation in the risk. After the data from the three states were removed, the CART tree selected is shown in \autoref{fig:finalcartmodel}. In this model, PS no longer appeared as a splitting variable. Otherwise, the structure of the tree remained similar to the previous CART model.


A previous study evaluated the effect of altitude on RSV-associated hospitalizations in Colorado from 1998 through 2002 and found that the risk for hospitalizations was much higher at elevations above 2500 meters \autocite{choudhuri2006}. For the three states considered here, the average elevation is approximately 6,800 feet for Colorado, 6,100 feet for Utah, and 5,700 feet for New Mexico. While high elevation was the observable characteristic of these three states, our analysis suggested that the underlying reason for the increased rates was lower pressure at these elevations, which resulted in less oxygen in the air. For infants and young children infected with RSV, reduced oxygen levels made their illnesses worse and increased the risk of hospitalization.


\subsubsection{Random Forest}
\begin{figure}[!htb]
	\begin{center}
	\includegraphics[width=1\linewidth]{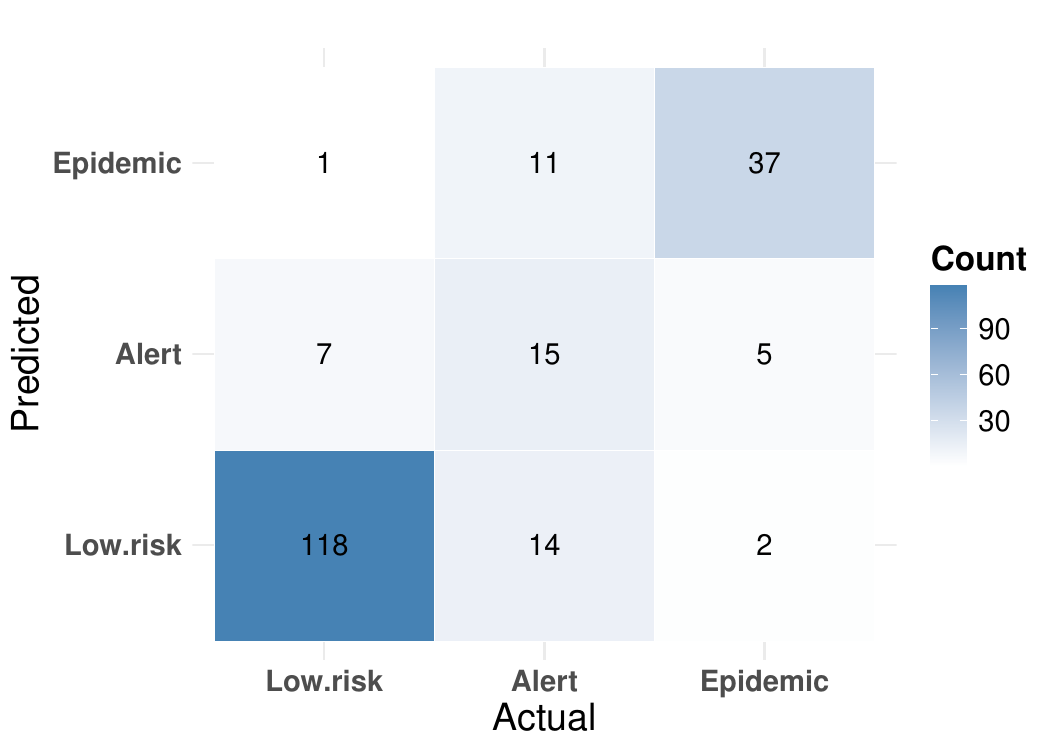} 
	\caption{Random Forest confusion matrix}
	\label{fig:confusionmatrixofrf}
    \end{center}
\end{figure}

\begin{figure}[!htb]
	 	\begin{center}
	 	\includegraphics[width=1\linewidth]{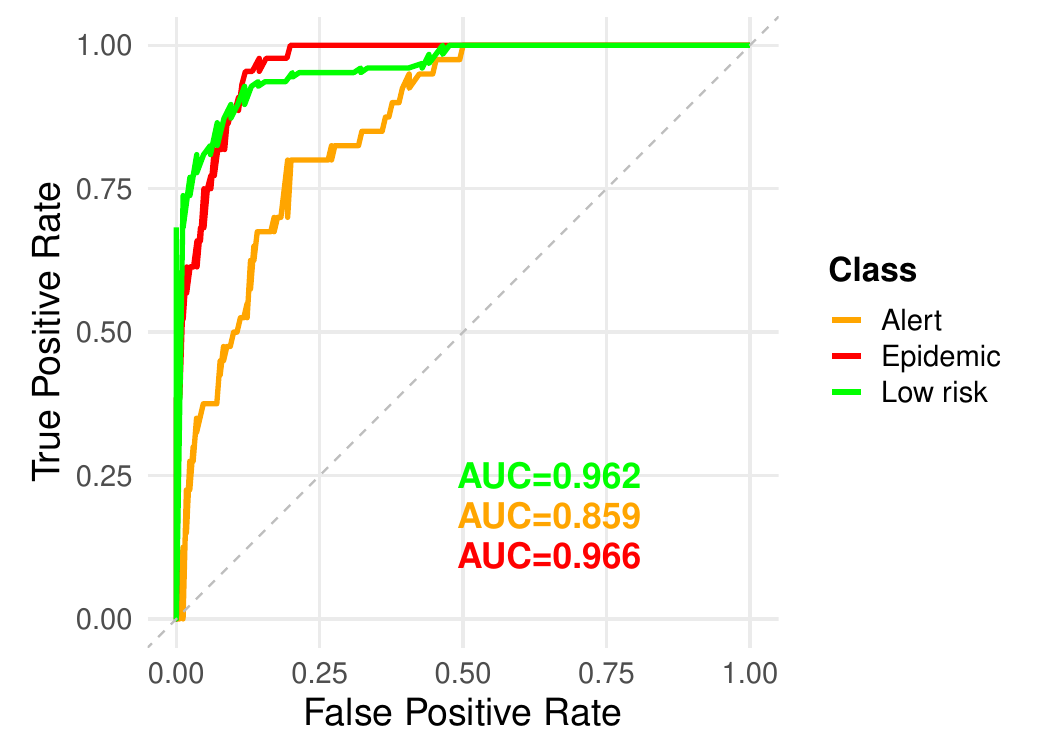} 
	 	\caption{Random Forest ROC}
	 	\label{fig:rocofrf}
	 	\end{center}
	 \end{figure}

Random forest is an extension of the decision tree algorithm, which can also be used to predict categorical response variables. We built a Random Forest classification model using the same training set with a 10-fold cross-validation. The model achieved a prediction accuracy of 0.810 with its confusion matrix shown in \autoref{fig:confusionmatrixofrf} and AUC values of 0.962, 0.859, and 0.966 for the three classes based on the ROC curves (\autoref{fig:rocofrf}), demonstrating strong prediction performance. The variable importance analysis indicated that \verb|WVAL| was the most important predictor, followed by \texttt{T2M, Ozone}, and \verb|QV2M|.

\subsubsection{Boosting}
\begin{figure}[!htb]
	\begin{center}
	\includegraphics[width=1\linewidth]{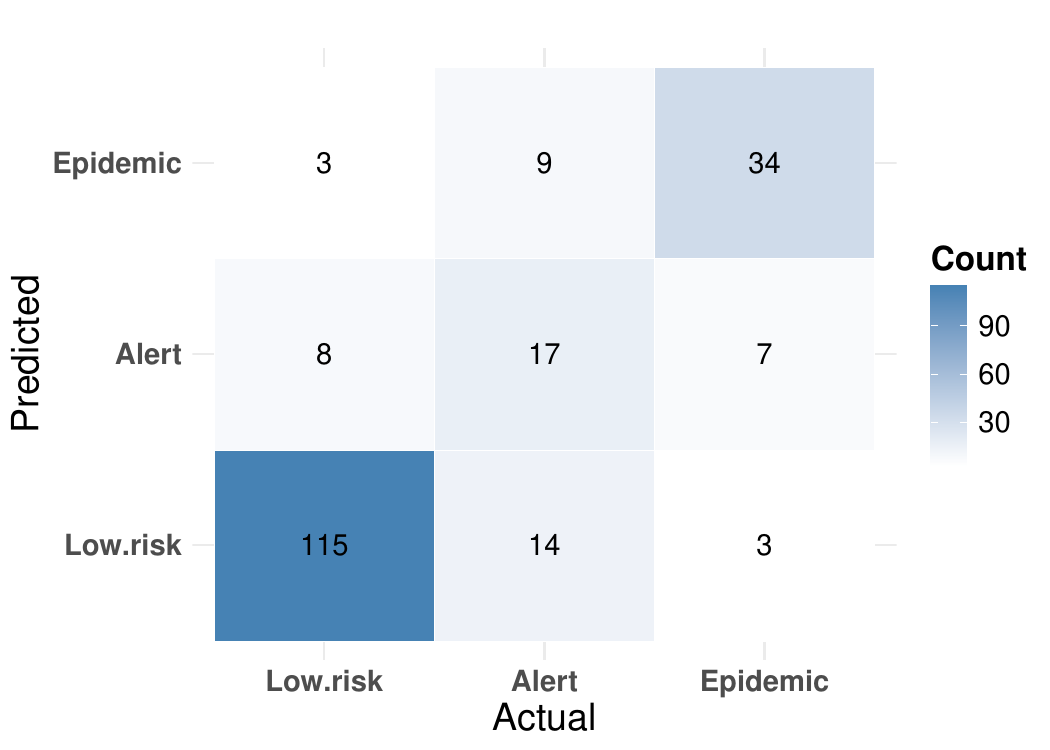} 
	\captionsetup{font=small}
	\caption{Boosting confusion matrix}
	\label{fig:confusionmatrixofbs}
	\end{center}
\end{figure}

\begin{figure}[!htb]
	 	\begin{center}
	 	\includegraphics[width=1\linewidth]{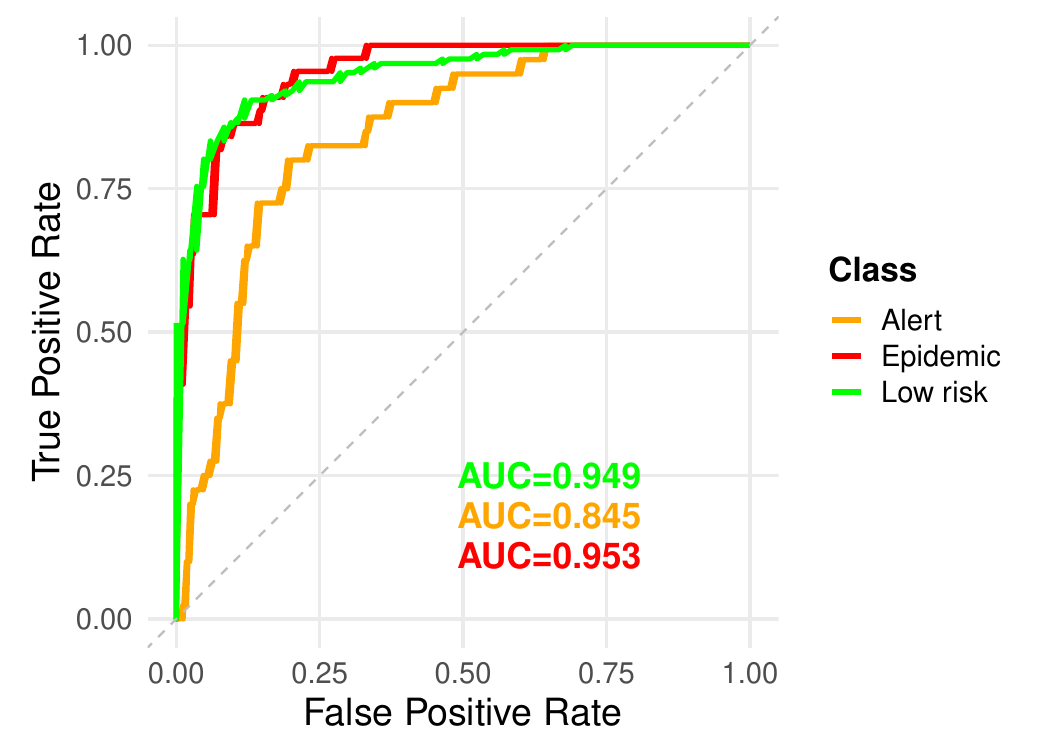} 
	 	\caption{Boosting ROC}
	 	\label{fig:rocofbs}
	 	\end{center}
	 \end{figure}

Boosting is an ensemble learning method that builds a series of decision trees sequentially to improve the previous ones. We applied a Boosting classification model to the same training set with a 10-fold cross-validation. The model achieved a prediction accuracy of 0.790 with its confusion matrix shown in \autoref{fig:confusionmatrixofbs} and AUC values of 0.949, 0.845, and 0.953 for the three classes based on the ROC curves (\autoref{fig:rocofbs}). The variable importance analysis showed that \verb|WVAL| was the top predictor, followed by \texttt{Ozone, PS}, and \verb|QV2M|, similar to the results for both CART and Random Forest models.

\subsubsection{Model Comparisons}
We compared the three models by applying them to the same testing set, evaluating their performance based on prediction accuracy, F-1 score, and AUC values (\autoref{tab:model_performance_classes}). Among the three models, Random Forest achieved the highest prediction accuracy, F-1 score, and AUC values across all three classes, indicating the best overall performance. Boosting showed slightly lower prediction accuracy, F-1 score, and AUC values, while CART ranked lowest among the three. The variable importance analyses were generally consistent across the models, with \verb|WVAL| identified as the most important predictor, followed by other variables such as \texttt{T2M, Ozone, QV2M}, or \verb|RSV Season|, although the order varied slightly.

\subsubsection{R Shiny Dashboard}
\begin{figure*}[!htb]
	\begin{center}
		\includegraphics[width=1\linewidth]{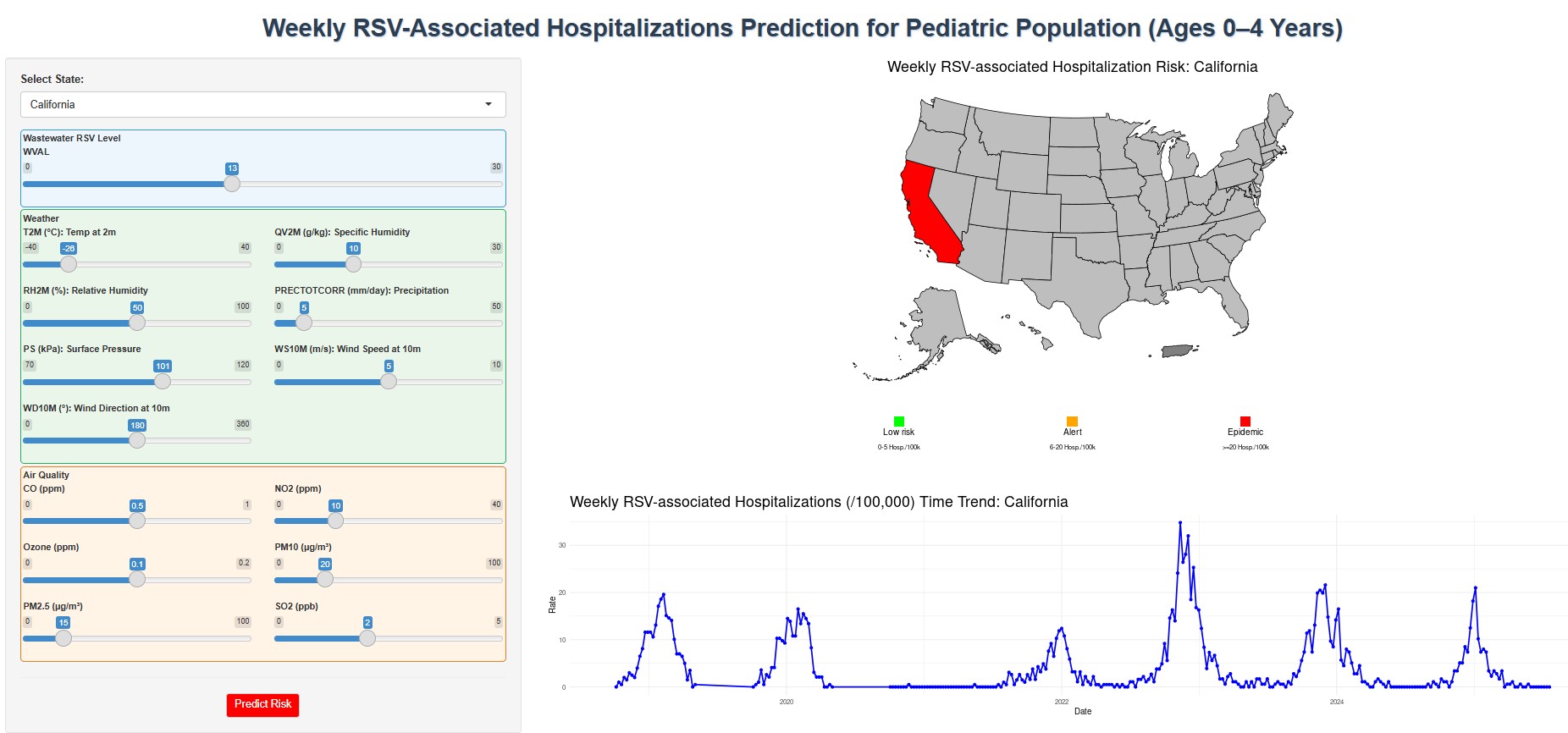} 
		\captionsetup{font=small}
		\caption{R Shiny dashboard}
		\label{fig:Rshiny}
	\end{center}
\end{figure*}

Based on the selected Random Forest model, we developed an \emph{R Shiny} dashboard (\autoref{fig:Rshiny}) at \url{https://f6yxlu-eric-guo.shinyapps.io/rsv_app/}, which allows users to predict the weekly rate of RSV-associated hospitalizations using environmental data. The dashboard provides interactive functionality where users can select a specific state and input wastewater RSV level \texttt{WVAL}, meteorological variables such as temperature, humidity, wind speed, etc, and air pollutant concentrations by using the sliding bar for each variable. The predicted risk will show up on the U.S. map. The dashboard provides a real-time risk prediction and visualizes the trends over time for the selected state based on the data available from RSV-NET, making it a practical tool for public health planning and early warning of potential hospitalization surges.

\section{Discussion}
Based on the selected Random Forest model, wastewater RSV level \verb|WVAL| ranked the first important predictor, which demonstrated the power of community wastewater surveillance in predicting the risk of RSV-associated hospitalizations. Air temperature at 2 meters above the ground (\verb|T2M|) was the second most important predictor, aligning with the seasonal pattern we observed for RSV-associated hospitalizations over the years. The ozone level was also a key predictor that contributed to the prediction, maybe due to its adverse effect on the human respiratory system, which increases the hospitalization rates of respiratory diseases \autocite{lu2023}. The fourth most important predictor was specific humidity at 2 meters above the ground (\verb|QV2M|). Previous studies have shown that higher specific humidity resulted in reduced RSV transmission \autocite{baker2019}. 

Our model was developed using the weekly rates for children aged 0--4 years, who are impacted most by RSV. However, the model framework can be applied to other age groups of interest as well. Our model was built upon wastewater RSV levels and other environmental data. Since influenza and SARS-CoV-2 viral activities are also monitored by wastewater surveillance. Our machine learning framework could be applied to predict hospitalizations associated with these pathogens as well. Currently, the application depends on datasets manually downloaded from the CDC and NASA databases, which limits its ability to provide real-time predictions automatically. Future work is to implement automated data integration so that the application updates as soon as new information becomes available from the meteorological, air quality, and wastewater surveillance databases. With the enhancements, this application would become an automated real-time monitoring system for multiple infectious disease-associated hospitalizations in the U.S. and help to monitor dynamic public health situations.

\phantomsection
\section*{Acknowledgments} 

\addcontentsline{toc}{section}{Acknowledgments} 

I would like to thank Dr. Kuruwita, Associate Professor of Statistics at Hamilton College, for his mentorship.  

\printbibliography

\end{document}